# Color Aesthetics and Social Networks in Complete Tang Poems: Explorations and Discoveries

**Chao-Lin Liu**[†] **Hongsu Wang**[‡] **Wen-Huei Cheng**[§] **Chu-Ting Hsu**[§] **Wei-Yun Chiu**[!]

[†§]Department of Computer Science, National Chengchi University, Taiwan

[‡]Institute for Quantitative Social Science, Harvard University, USA

[§ !]Department of Chinese Literature, National Chengchi University, Taiwan

[†]Graduate Institute of Linguistics, National Chengchi University, Taiwan

{[†]chaolin,[§]104753021,[§]whcheng}@nccu.edu.tw, [‡]hongsuwang@fas.harvard.edu, [ǂ]acwu0523@gmail.com

## Abstract[1]

The *Complete Tang Poems* (CTP) is the most important source to study Tang poems. We look into CTP with computational tools from specific linguistic perspectives, including distributional semantics and collocational analysis. From such quantitative viewpoints, we compare the usage of “wind” and “moon” in the poems of Li Bai[2] (李白) and Du Fu (杜甫). Colors in poems function like sounds in movies, and play a crucial role in the imageries of poems. Thus, words for colors are studied, and “白” (bai2, white) is the main focus because it is the most frequent color in CTP. We also explore some cases of using colored words in antithesis(對仗)[3] pairs that were central for fostering the imageries of the poems. CTP also contains useful historical information, and we extract person names in CTP to study the social networks of the Tang poets. Such information can then be integrated with the China Biographical Database of Harvard University.

## 1 Introduction

*Complete Tang Poems* (CTP) is the single most important collection for studying Tang poems from the literary and linguistic perspectives (Fang et al., 2009; Lee and Wong, 2012). CTP was officially compiled during the Kangxi years of the Qing dynasty, and includes more than 40,000 poems, totaling more than 3 million characters, that were authored by more than 2000 poets. Employing linguistic theories and computational tools, we analyze the contents of CTP for a wide variety of explorations.

Lo and her colleagues pioneered to handle texts of Chinese classical poetry with computer software (Lo et al., 1997). Hu and Yu (2001) achieved a better environment and demonstrated its functions with a temporal analysis of selected Chinese unigrams, i.e., 愁(chou2), 苦(ku3), 恨(hen4), 悲(bei1), 哀(ai1), and 憂(you1). Jiang (2003) employed tools for information retrieval to find and study selected poems of Li Bai and Du Fu that mentioned “wind” and “moon”. Huang (2004) analyzed the ontology in Su Shi’s (蘇軾) poems based on 300 Tang Poems, and Chang et al. (2005) continued this line of work. Lo then built a more complete taxonomy for Tang poems (Lo, 2008; Fang et al., 2009).

Lee conduct part-of-speech analysis of CTP (Lee, 2012) and dependency trees (Lee and Kong, 2012). They also explored the roles of a variety of named entities, e.g., seasons, directions, and colors, in CTP (Lee and Wong, 2012), and used their analysis of CTP for teaching computational linguistics (Lee et al., 2013).

CTP can serve as the bases of other innovative applications. Zhao and his colleagues have created a website[4] for suggesting couplets, which was accomplished partially based on their analysis of the CTP (Jiang and Zhou 2008; Zhou et al., 2009). Voigt and Jarafsky (2013) considered CTP when they compared ancient and modern verses of China and Taiwan.

---

[1] A Chinese version of this paper appeared in the *Proc. of the 27th Conference on Computational Linguistics and Speech Analysis* (Liu et al., 2015).

[2] Romanized Chinese names are in the order of surname and first name, following the request of a reviewer.

[3] “Antithesis” is not a perfect translation of “對仗” (dui4 zhang4). Roughly speaking, “對仗” refers to constrained collocations, and requires two terms to have opposite relationships in pronunciations, but does not demand the terms to be opposite in meanings. In English, “antithesis” carries a rather obvious demand for two referred terms to be opposite in meanings.

[4] http://couplet.msra.cn/ of Microsoft Research Asia

Our work is special in that we analyze the contents of CTP from some linguistic perspectives, including collocational analysis and distributional semantics. We rely on certain literary knowledge for handling the CTP texts such that typical procedures for text analysis are not absolutely necessary for obtaining the words in poems. The concept of distributional semantics (Harris, 1954; Miller and Walter, 1991) provides the basis for comparing poets' styles. We investigate the functions of colors in poems by considering their collocations and antitheses. Colors in poems play similar roles as sounds and lights in movies. They are crucial for nurturing the imageries in art works. Analyzing the appearances of colors in poems leads to interesting observations.

We extend our exploration from literature to history in CTP, just like Chen's (2010) studies of the political information hidden in the poems of Tang Taizong (唐太宗). Person names that were mentioned in the poems provide hints about the social networks of the poets. Hence, we may employ CTP as a source of biographical information for the China Biographical Database of Harvard University.

In Section 2, we check the CTP used in our work, and report a basic analysis of its contents. In Section 3, through a distributional semantics perspective, we compare the usage of "wind" and "moon" of Li Bai and Du Fu, and examine why some poets, e.g., Bai Juyi (白居易), were classified as a social poet (社會詩人, she4 hui4 shi1 ren2). In Section 4, we dig into the usage of more colors and related words by considering their collocations and antitheses. In Section 5, we show and discuss how social networks of poets can be computed from CTP. In Section 6, we extend applications and analyses of CTP to couplet suggestion and authorship attribution.

## 2 Data Sources and Basic Analysis

Although we have found several papers on the analyses of CTP, none of them specified exactly which version of CTPs was used in their work. Although there is only version of CTP in "欽定四庫全書"[5] (qin1 ding4 si4 ku4 quan2 shu1), there exist alternative text versions.

### 2.1 Data Sources

There is no "the" authoritative version of CTP. In "御定全唐詩" (yu4 ding4 quan2 tang2 shi1), we can find Li Bai's "牀前看月光"[6] (chuang2 qian2 kan4 yue4 guang1), but, in textbooks in Taiwan, we will read "牀前明月光" (chuang2 qian2 ming2 yue4 guang1). Both are accepted by domain experts.

We can find online CTP in WikiSource[7], Wenxue100[8], Xysa[9], Ctext[10], ChillySpring[11], and Guji[12], for example. Most of these websites allow online reading but do not allow complete download, though some do. We have completed a preliminary comparison between the Wenxue100 and Ctext versions. They are very similar, and appear to have a common source.

In this paper we are using the version that we obtained from Wenxue100.

### 2.2 Basic Analyses

In the Wenxue100 version, we have 42,863 works, and, in Figure 1, we show the poets who have leading numbers of works in CTP. The vertical axis shows the numbers of their works included in CTP, and poets' Chinese names are shown along the horizontal axis, where "不詳" (bu4 xiang2) means unknown and is not a name. Bai Juyi is the most popular poet in CTP, and has 2643 works, followed by Du Fu's 1158 works and Li Bai's 896 works. In total, we have 3,055,044 Chinese characters and punctuations in this version of CTP.

---

[5] https://en.wikipedia.org/wiki/Siku_Quanshu
[6] source: Li Bai (李白): 靜夜思 (jing4 ye4 si1)
[7] WikiSource: https://zh.wikisource.org/zh-hant/
[8] Wenxue100: http://www.wenxue100.com
[9] Xysa: http://www.xysa.com/
[10] Ctext: http://www.ctext.org/
[11] ChillySpring: http://210.69.170.100/s25/
[12] Guji: http://guji.artx.cn/

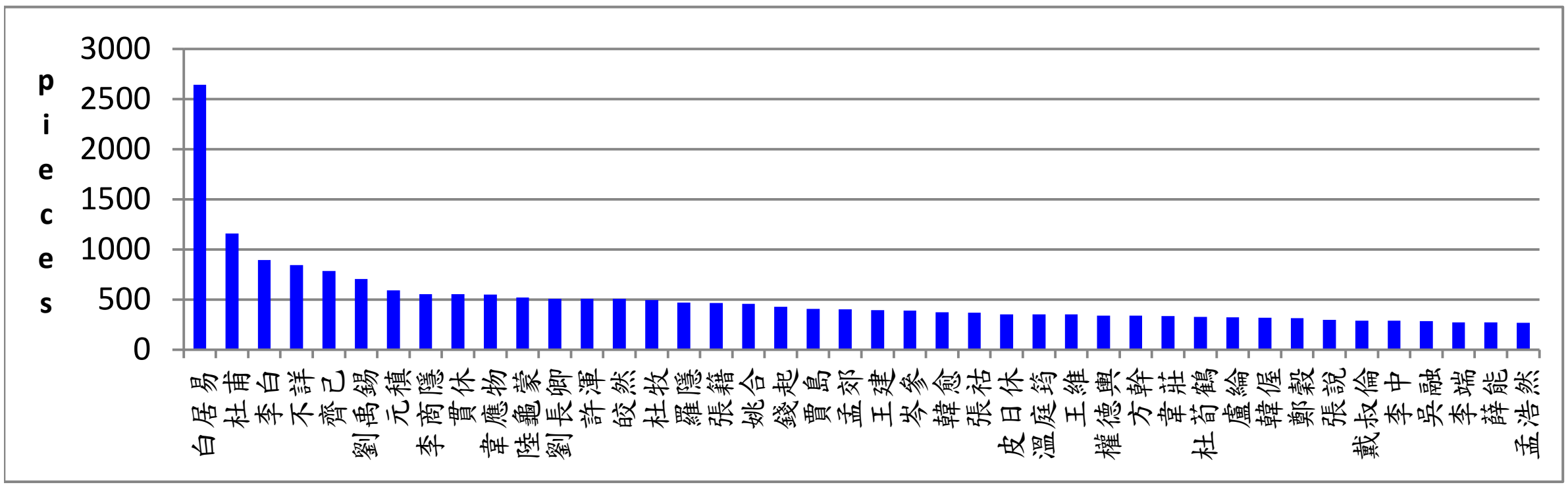

**Figure 1. Number of works of the leading poets in CTP**

The exact number of works in CTP needs further research to make sure. Some adjustments should be expected. We cannot determine the authors of some poems even when we read the "御定全唐詩". The author of the poem with title "題霍山秦尊師" (ti2 huo4 shan1 qin2 zun1 shi1) may be Du Gaung-Ting (杜光庭) or Zheng Ao (鄭遨) as the piece appeared in volumes 854 and 855, respectively.

It is very easy to compute the most common unigrams in CTP, and they are "不"(bu4), "人"(ren2), "山"(shan1), "無"(wu2), "風"(feng1), "一"(yi1), "日"(ri4), "雲"(yun2), "有"(you3), and "何"(he2).

It is very challenging to precisely segment words in poems without human final inspection. Nevertheless, it is well-known that the patterns of 5-character and 7-character Tang poems usually follow specific traditions (Lo, 2005). The segments in poems were constituted by words of one or two characters.

For 5-character Tang poems, the sentences in poems can be segmented into words of 2, 2, and 1 characters or alternatively 2, 1, 2 characters. For instance "白日依山盡"[13] (bai2 ri4 yi1 shan1 jin4) and "感時花濺淚"[14] (gan3 shi2 hua1 jian4 lei4) used, respectively, 2+2+1 and 2+1+2 patterns. Similarly, the sentences of 7-character Tang poems usually used 2+2+2+1 and 2+2+1+2 patterns, e.g., "東風不與周郎便"[15] (dong1 feng1 bu4 yu3 zhou1 lang2 bian4) and "晉代衣冠成古丘"[16] (jin4 dai4 yi1 guan1 cheng2 gu3 qiu1), respectively.

Employing such a literary common sense as a heuristic for segmenting words in CTP, we can find words of relatively high frequencies in Table 1. In the order of higher frequencies, the most common two-character words in CTP are "何處" (he1 chu4, where), "不知" (bu4 zhi1, unknown), "萬里" (wan4 li3, tens of thousands of miles), "千里" (qian1 li3, thousands of miles), "今日" (jin1 ri4, today), "不見" (bu2 jian4, cannot be seen), "不可" (bu4 ke3, cannot), "春風" (chun1 feng1, spring wind), "白雲" (bai2 yun2, white cloud), "不得"(bu4 de2, cannot), "明月" (ming2 yue4, bright moon) 和"人間" (ren2 jian1, human world).

Not all frequent strings thus identified are real words. "人不" (ren2 bu4) in Table 1 is not a word but just a frequent string as in "盡日傷心人不見"[17] (jin4 ri4 shang1 xin1 ren2 bu2 jian4) and "雖病人不知"[18] (sui1 bing4 ren2 bu4 zhi1).

Proposing nonwords like "人不" brings researchers inconvenience but does not cause serious troubles. Considering that we are handling millions of characters in CTP and that it is neither impossible nor very time-consuming to identify such nonwords, our experience indicates that applying computational tools for automatic processing of the CTP contents made our study of CTP more efficient than without using the tools.

---

13 source: Wang Zhi-Huan (王之渙): 登鸛雀樓 (deng1 guan4 que4 lou2)
14 source: Du Fu (杜甫): 春望 (chun1 wang4)
15 source: Li Bai (李白): 登金陵鳳凰臺(deng1 jin1 ling2 feng4 huang2 tai2)
16 source: Du Mu (杜牧): 赤壁 (chi4 bi4)
17 source: Li Shang-Yin (李商隱): 遊靈伽寺(you2 ling2 qie2 si4)
18 source: Bai Juyi (白居易): 讀史五首(du2 shi3 wu3 shou3)

**Table 1. Frequent bigrams in CTP**

| bigram | freq. | bigram | freq. | bigram | freq. | bigram | freq. | bigram | freq. |
|---|---|---|---|---|---|---|---|---|---|
| 何處 | 1669 | 無人 | 881 | 青山 | 662 | 流水 | 550 | 落日 | 498 |
| 不知 | 1469 | 風吹 | 834 | 少年 | 634 | 回首 | 544 | 不如 | 497 |
| 萬里 | 1455 | 惆悵 | 780 | 相逢 | 629 | 可憐 | 539 | 歸去 | 496 |
| 千里 | 1305 | 故人 | 778 | 平生 | 597 | 如此 | 526 | 日暮 | 496 |
| 今日 | 1165 | 秋風 | 749 | 年年 | 593 | 白髮 | 520 | 不能 | 481 |
| 不見 | 1158 | 悠悠 | 740 | 寂寞 | 592 | 主人 | 517 | 別離 | 481 |
| 不可 | 1148 | 相思 | 733 | 黃金 | 589 | 今朝 | 516 | 何時 | 478 |
| 春風 | 1128 | 長安 | 722 | 天子 | 588 | 月明 | 515 | 此時 | 477 |
| 白雲 | 1108 | 白日 | 697 | 人不 | 587 | 從此 | 509 | 洛陽 | 476 |
| 不得 | 947 | 如何 | 687 | 天地 | 586 | 日月 | 508 | 天下 | 472 |
| 明月 | 896 | 十年 | 678 | 何事 | 579 | 行人 | 507 | 芳草 | 472 |
| 人間 | 890 | 何人 | 663 | 江上 | 553 | 將軍 | 499 | 歸來 | 471 |

We have designed tools for extracting contexts that contain candidate words for researchers' inspections, so validating candidate words is very easy. Furthermore, sometimes the words that do not exist in contemporary literature may turn out to be artistic usages of words that are hard to be correctly handled by current tools for Chinese segmentation, and researchers are more than happy to check those innovative words with a limited cost in time.

Table 1 exemplifies a problem of this paper: that we cannot provide pronunciations for all Chinese words that appear in this page-limited manuscript. Due to the large number of Chinese words in this and other tables, we cannot afford to annotate and explain all of the words in tables.

## 3 Styles

Adapting the concept of distributional semantics[19] (Harris, 1954; Miller and Walter 1991), researchers investigate the semantics of a word from the distributions of its surrounding words. Firth (1957) stated that "**You shall know a word by the company it keeps.**" Similarly, we should be able to extend the concept of distributional semantics to compare poets' styles: "**You shall know a poet's style by the words s/he uses**."

### 3.1 Wind and Moon

Jiang (2003) compared the styles of Li Bai and Du Fu by looking into how they used "風" (feng1, wind) and "月" (yue4, moon) in their works by checking into individual pieces of poems. We take a quantitative approach by examining the terms related to "風" and "月" in the poets' works.

We can employ the PAT-tree techniques (Chien, 1997) or our own tools[20] for finding the characters that appear immediately before a specified Chinese character to find words containing "風" and "月" in poets' works.

Tables 2 and 3 show how Li and Du used "wind." Both Li and Du were very creative in inventing terms for "風". The most common term of "風" in Li Bai's works were "春風" (chun1 feng1, spring wind), "清風" (qing1 feng1, clear wind), "秋風"(qiu1 feng1, autumn wind), "東風" (dong1 feng1, eastern wind), and "長風" (zhang2 feng1, long wind). Du Fu , on the other hand, had "秋風", "春風", "北風" (bei3 feng1, northern wind), "悲風" (bei1 feng1, sad wind), and "朔風" (shuo4 feng1, northern wind).

---

[19] A reviewer for this paper, referring to (Lin, 1998), considers our approach as a collocational analysis of words in a given context. We used "distributional" because there is a sense of distribution when we can consider the frequency distribution of a set of words.

[20] https://sites.google.com/site/taiwandigitalhumanities/

**Table 2. Li Bai's wind**

| bigram | freq. | bigram | freq. | bigram | freq. | bigram | freq. | bigram | freq. |
|---|---|---|---|---|---|---|---|---|---|
| 春風 | 72 | 松風 | 17 | 南風 | 8 | 悲風 | 6 | 高風 | 4 |
| 清風 | 28 | 隨風 | 14 | 北風 | 8 | 飄風 | 5 | 西風 | 4 |
| 秋風 | 26 | 香風 | 11 | 涼風 | 8 | 胡風 | 5 | 扶風 | 4 |
| 東風 | 24 | 天風 | 10 | 狂風 | 7 | 從風 | 5 | 屏風 | 4 |
| 長風 | 22 | 英風 | 8 | 雄風 | 6 | 嚴風 | 5 | 動風 | 4 |

**Table 3. Du Fu's wind**

| bigram | freq. | bigram | freq. | bigram | freq. | bigram | freq. | bigram | freq. |
|---|---|---|---|---|---|---|---|---|---|
| 秋風 | 30 | 朔風 | 8 | 高風 | 6 | 江風 | 4 | 南風 | 4 |
| 春風 | 19 | 微風 | 8 | 清風 | 6 | 驚風 | 4 | 涼風 | 4 |
| 北風 | 14 | 隨風 | 7 | 天風 | 6 | 山風 | 4 | 東風 | 4 |
| 悲風 | 10 | 回風 | 7 | 長風 | 5 | 多風 | 4 | | |
| 裡風 | 8 | 臨風 | 7 | 陰風 | 4 | 含風 | 4 | | |

**Table 4. Li Bai's moon**

| bigram | freq. | bigram | freq. | bigram | freq. | bigram | freq. | bigram | freq. |
|---|---|---|---|---|---|---|---|---|---|
| 明月 | 57 | 溪月 | 9 | 有月 | 5 | 湖月 | 3 | 夜月 | 3 |
| 秋月 | 40 | 八月 | 9 | 轉月 | 4 | 漢月 | 3 | 夕月 | 3 |
| 五月 | 28 | 雲月 | 9 | 曉月 | 4 | 樓月 | 3 | 端月 | 3 |
| 日月 | 23 | 花月 | 8 | 孤月 | 4 | 新月 | 3 | 向月 | 3 |
| 海月 | 14 | 見月 | 7 | 台月 | 4 | 待月 | 3 | 古月 | 3 |
| 上月 | 13 | 江月 | 6 | 落月 | 3 | 弄月 | 3 | 十月 | 3 |
| 三月 | 13 | 蘿月 | 5 | 片月 | 3 | 如月 | 3 | 二月 | 3 |
| 山月 | 10 | 素月 | 5 | 滿月 | 3 | 好月 | 3 | 乘月 | 3 |

**Table 5. Du Fu's moon**

| bigram | freq. | bigram | freq. | bigram | freq. | bigram | freq. | bigram | freq. |
|---|---|---|---|---|---|---|---|---|---|
| 日月 | 20 | 明月 | 7 | 落月 | 4 | 正月 | 3 | 從月 | 3 |
| 歲月 | 14 | 江月 | 6 | 秋月 | 4 | 星月 | 3 | 九月 | 3 |
| 十月 | 10 | 五月 | 6 | 漢月 | 4 | 新月 | 3 | | |
| 三月 | 9 | 夜月 | 5 | 門月 | 3 | 四月 | 3 | | |
| 八月 | 8 | 二月 | 5 | 素月 | 3 | 六月 | 3 | | |

For those who can appreciate Chinese, the poets' "風" carried very different imageries. For instance, in China, winds from the north are generally cold, which has been used to convey a sense of sadness by many. In contrast, the eastern wind and spring wind are more comfortable and pleasing.

Tables 4 and 5 show how Li and Du used "moon." Although Du had more works in CTP (cf. Section 2.2), statistics in Tables 4 and 5 show that Li used more words about "月" than Du. Li and Du also demonstrated quite diverging styles in using "月". For frequent terms of "月", Li had "明月" (ming2 yue4, bright moon), "秋月" (qiu1 yue4, autumn moon), "日月" (ri4 yue4, sun and moon) , "海月" (hai3 yue4, sea and moon), and "山月"(shan1 yue4, mountain moon); in contrast, Du used more "歲月" (sui4 yue4, ages) and monthly names, e.g., "十月" (shi2 yue4, October), "三月" (san1 yue4, March), and "八月" (ba1 yue4, August).

## 3.2 White Words in CTP

Colors for poems are like sounds for movies. They foster the feelings and imageries of the artistic works. When we checked the most frequent unigrams in CTP (cf. Section 2.2), we have found that "白" (bai2, white) is the most frequent color name in CTP.

Using the same mechanism for finding words that contained "風" and "月" in CTP, we can find words that started with "白". Then, we can calculate the percentage of a poet's poems that used spe-cific words that started with "白". The statistics are collected for 13 renowned poets and are listed in Table 6, which is placed at the end of this manuscript because of its huge size.

**Table 6. Percentages of poets' works that used white words**

| Ratio A | | 8.96 | 18.41 | 9.73 | 46.65 | 23.83 | 12.55 | 26.94 | 15.67 | 18.80 | 17.37 | 16.30 | 10.48 |
|---|---|---|---|---|---|---|---|---|---|---|---|---|---|
| Ratio B | | 1.87 | 5.72 | 1.80 | 5.92 | 2.13 | 4.66 | 7.94 | 1.99 | 2.28 | 7.19 | 3.70 | 3.23 |
| **freq.** | **bigram** | **孟浩然** | **孟郊** | **李商隱** | **李白** | **李賀** | **杜牧** | **杜甫** | **溫庭筠** | **王維** | **白居易** | **賈島** | **韓愈** |
| 217 | 白日 | 0.75 | 4.73 | 1.62 | 6.92 | 2.98 | 1.01 | 2.42 | 0.00 | 1.14 | 2.04 | 2.22 | 3.23 |
| 164 | 白髮 | 1.12 | 3.73 | 0.54 | 2.34 | 1.28 | 1.62 | 1.99 | 0.00 | 0.85 | 2.50 | 2.22 | 0.54 |
| 158 | 白雲 | 2.99 | 1.99 | 0.54 | 3.79 | 0.85 | 1.42 | 0.86 | 0.28 | 7.41 | 0.95 | 4.44 | 0.27 |
| 149 | 白頭 | 0.00 | 0.75 | 0.72 | 0.67 | 0.43 | 2.23 | 3.37 | 1.14 | 0.57 | 2.23 | 0.49 | 1.61 |
| 86 | 白首 | 0.75 | 1.00 | 0.18 | 1.56 | 0.43 | 0.20 | 1.99 | 0.85 | 0.85 | 1.02 | 0.25 | 0.81 |
| 74 | 白玉 | 0.00 | 0.50 | 2.34 | 3.01 | 0.85 | 0.81 | 0.60 | 0.00 | 0.85 | 0.53 | 0.00 | 0.27 |
| 74 | 白馬 | 0.37 | 0.50 | 0.00 | 2.34 | 4.68 | 0.00 | 1.38 | 2.85 | 0.85 | 0.30 | 0.00 | 0.00 |
| 63 | 白雪 | 0.37 | 0.25 | 0.36 | 2.34 | 0.00 | 0.40 | 1.04 | 0.28 | 0.00 | 0.68 | 0.00 | 0.27 |
| 59 | 白帝 | 0.00 | 0.00 | 0.18 | 1.00 | 0.43 | 0.00 | 3.54 | 0.28 | 0.00 | 0.08 | 0.00 | 0.54 |
| 58 | 白露 | 0.00 | 0.50 | 0.18 | 1.56 | 0.43 | 0.00 | 0.86 | 0.28 | 0.28 | 0.79 | 0.99 | 0.27 |
| 54 | 白石 | 0.00 | 1.00 | 0.90 | 1.12 | 0.43 | 0.00 | 0.26 | 0.57 | 0.57 | 0.68 | 1.23 | 0.81 |
| 38 | 白蘋 | 0.37 | 0.75 | 0.18 | 0.22 | 1.28 | 0.20 | 0.52 | 2.85 | 0.00 | 0.30 | 0.00 | 0.54 |
| 32 | 白水 | 0.00 | 0.25 | 0.00 | 0.89 | 1.28 | 0.00 | 1.12 | 0.00 | 0.57 | 0.08 | 0.00 | 0.27 |
| 31 | 白鬚 | 0.00 | 0.00 | 0.18 | 0.11 | 0.00 | 0.20 | 0.00 | 0.00 | 0.00 | 0.98 | 0.49 | 0.00 |
| 30 | 白鷺 | 0.00 | 0.25 | 0.00 | 1.79 | 0.00 | 0.40 | 0.26 | 0.00 | 0.85 | 0.15 | 0.25 | 0.00 |
| 21 | 白骨 | 0.00 | 0.25 | 0.18 | 1.23 | 0.00 | 0.00 | 0.60 | 0.00 | 0.00 | 0.00 | 0.00 | 0.27 |
| 17 | 白衣 | 0.00 | 0.00 | 0.18 | 0.22 | 0.00 | 0.00 | 0.17 | 0.00 | 0.57 | 0.19 | 0.99 | 0.00 |
| 15 | 白髭 | 0.00 | 0.00 | 0.00 | 0.00 | 0.00 | 0.40 | 0.00 | 0.00 | 0.00 | 0.45 | 0.25 | 0.00 |
| 15 | 皓齒 | 0.00 | 0.00 | 0.00 | 0.78 | 0.85 | 0.00 | 0.26 | 0.85 | 0.00 | 0.00 | 0.00 | 0.00 |

The third and following rows of Table 6 show two types of information. The second leftmost column lists the white words that appeared more than 30 times in the works of 12 poets[21]. The left-most column lists the total frequencies of these white words that were used in the poets' poems. The percentages that appear to the right of the white words indicate how often an individual poet used this white word, while the thick boxes indicate the most frequent words used by the poets.

The first row, Ratio A, of Table 6 shows the total percentage of the poet's poems in CTP that used the white words in Table 6. Li Bai liked to use "白" much more than others based on the data. Nearly half of Li's poems in CTP, i.e., 46.65%, used the color.

The second row, Ratio B, of Table 6 shows the total percentage of poets' poems that used "白髮" (bai2 fa3, gray hair), "白頭"(bai2 tao2, white head), "白首" (bai2 shou3, white head), "白鬚"(bai2 syu1, white beard), "白骨"(bai2 gu3, white bone), and "白髭" (bai2 zi1, white mustache). Statistics for these terms are specially marked by shadowed rows. These six terms typically appeared in works that carried pessimistic senses.

It is thus possible to peek into the differences of the main themes of poets' works with Ratio B. The B ratios of Meng Hao-Ran (孟浩然), Li Shang-Yin (李商隱), and Wen Ting-Yun (溫庭筠) are less than 2%. In sharp contrast, we could see that the B ratios of Du Fu and Bai Juyi are more than 7%. In general, Meng's works are considered to belong to the relaxing category (田園詩派, Tian2 Yuan2 Shi1 Pai4), and Li and Wen are considered to use expressions that lead to "a beautiful and gorgeous conception" (Lee, 2009). Both Du Fu and Bai Juyi, on the other hand, are considered as social poets who cared about the status of the society.

---

[21] Table 6 also includes statistics for selected words that appeared more less than 30 times. This table is adapted from a similar table in (Liu et al., 2015).

**Table 7. Statistics of some collocations in CTP (*n*=30)**

| 白雲 | | | | 白日 | | 白髮 | | | |
|---|---|---|---|---|---|---|---|---|---|
| bigram | freq. | bigram | freq. | bigram | freq. | bigram | freq. | bigram | freq. |
| 明月 | 61 | 清露 | 10 | 青春 | 32 | 青山 | 38 | 丹砂 | 7 |
| 流水 | 40 | 青壁 | 7 | 青山 | 21 | 青雲 | 27 | 黃河 | 6 |
| 芳草 | 29 | 秋草 | 7 | 清風 | 18 | 朱顏 | 16 | 清光 | 4 |
| 滄海 | 28 | 丹灶 | 5 | 紅塵 | 15 | 青春 | 15 | 丹霄 | 4 |
| 紅葉 | 17 | 青鏡 | 2 | 黃河 | 15 | 黃金 | 13 | 黃衣 | 3 |
| 黃葉 | 16 | 青玉 | 2 | 滄江 | 6 | 滄洲 | 8 | 紅塵 | 3 |
| 青草 | 14 | 皇道 | 1 | 青蓮 | 3 | 青衫 | 7 | 紅旗 | 3 |
| | | | | 青霄 | 3 | | | | |
| | | | | 青楓 | 2 | | | | |

## 4 Collocations and Antithesis

Collocations refer to the occurrence of two words within a specified range. If two words constantly appear within a short range, they may have some close relationships in semantics.

Antithesis (對仗, dui4 zhang4) refers to the parallelism in poems. Two words need to have special relationships in their positions, pronunciations, and meanings to form an antithesis pair.

There are complex rules[22] for observing antitheses and rhymes in Tang poems. In a Lu Shi (律詩), the third and the fourth sentences form a sentence pair (聯, lian2), so do the fifth and the sixth sentences. A pair of sentences should follow the antithesis rules. In "白日當空天氣暖，好風飄樹柳陰涼"[23], "白日" and "柳陰" form a collocation but not antithesis, while "白日" and "好風" are both a pair of collocation and a pair of antithesis. In this case, "白日" and "柳陰" do not locate at corresponding positions.

If we can segment words in poems correctly, then recognizing antithesis will not be very difficult. However, perfect word segmentation in poems needs semantic information. Sometimes the poems could carry ambiguous meanings.

### 4.1 Word Pairs

We can compute the collocations of a word to acquire a sense of the circumstances of the word's occurrences. To do so, we extract contexts of words, say ***n*** characters, before and after the word of interest from CTP. Then we can compute frequent words from the contexts (cf. Section 2.2).

Table 7 lists some educational findings when we set ***n*** to 30. It is very interesting to find out that "白雲" (bai2 yun2, white cloud) collocates with "明月" (ming2 yue4, bright moon) and "流水" (liu2 shui3, running water), that "白日" (bai2 ri4, bright sun) collocates with "青春" (qing1 chun1, young age) and "青山" (qing1 shan1, mountains), and that "白髮" (bai2 fa3, gray hair) collocates with "青山" and "青雲" (qing1 yun2, blueish clouds).

We can also study the cases of antitheses of the white words that were used by individual poets. For instance, we can find at least 26 instances of "白髮" and "青雲" that were used as a antithesis pair in CTP. "白雲" and "流水" were used as a antithesis pair by Liu Yu-Xi (劉禹錫), Yao He (姚合), Huang-fu Ran (皇甫冉), Huang-fu Cent (皇甫曾), Jia Dao (賈島), and Qian Qi (錢起), while "白雲" and "青草" (qing1 cao3, green grass) were used as a antithesis pair by Liu Zhang-Qing (劉長卿), Shu-kong Si (司空曙), Yao Ho (姚合), Zhang Ji (張籍), Li Tuan (李端), and Lang Shi-Yuan (郎士元).

These statistics offer some hints about the word semantics, and researchers may want to (and we can) extract the poems that contain a specific pair of words to examine the complete poems for either literary or social studies.

---

22 http://cls.hs.yzu.edu.tw/300/all/primary1/DET4.htm

23 source: Yuan Zhen (元稹): 清都春霽，寄胡三、吳十一 (qing1 dou1 chun1 ji4，ji4 hu2 san1、wu2 shi2 yi1)

**Table 8. Corresponding colors in CTP**

| 白 | | 青 | | 紅 | | 黃 | | 綠 | | 紫 | | 碧 | | 丹 | | 赤 | | 黑 | |
|---|---|---|---|---|---|---|---|---|---|---|---|---|---|---|---|---|---|---|---|
| **C** | **F** | **C** | **F** | **C** | **F** | **C** | **F** | **C** | **F** | **C** | **F** | **C** | **F** | **C** | **F** | **C** | **F** | **C** | **F** |
| 青 | 919 | 白 | 919 | 白 | 358 | 白 | 505 | 紅 | 335 | 青 | 197 | 紅 | 199 | 白 | 142 | 青 | 54 | 青 | 36 |
| 黃 | 505 | 綠 | 202 | 綠 | 335 | 青 | 152 | 青 | 202 | 黃 | 139 | 青 | 188 | 紫 | 70 | 黃 | 39 | 黃 | 27 |
| 紅 | 358 | 紫 | 197 | 碧 | 199 | 紫 | 139 | 黃 | 83 | 紅 | 107 | 清 | 100 | 碧 | 50 | 白 | 33 | 紅 | 24 |
| 清 | 274 | 碧 | 188 | 翠 | 139 | 綠 | 83 | 白 | 70 | 白 | 72 | 黃 | 74 | 青 | 41 | 紫 | 19 | 白 | 15 |
| 丹 | 142 | 黃 | 152 | 青 | 111 | 碧 | 74 | 清 | 70 | 丹 | 70 | 白 | 57 | 翠 | 35 | 蒼 | 15 | 明 | 13 |
| 滄 | 99 | 紅 | 111 | 紫 | 107 | 紅 | 44 | 丹 | 31 | 清 | 56 | 丹 | 50 | 綠 | 31 | 紅 | 13 | 清 | 10 |
| 朱 | 97 | 翠 | 54 | 黃 | 44 | 赤 | 39 | 朱 | 27 | 朱 | 41 | 金 | 42 | 玉 | 29 | 滄 | 12 | 丹 | 8 |
| 明 | 96 | 赤 | 54 | 清 | 36 | 翠 | 33 | 紫 | 26 | 金 | 39 | 紫 | 35 | 素 | 25 | 丹 | 10 | 寒 | 8 |
| 綠 | 70 | 明 | 42 | 素 | 31 | 清 | 32 | 碧 | 26 | 碧 | 35 | 朱 | 31 | 金 | 21 | 清 | 8 | 紫 | 7 |
| 玄 | 66 | 丹 | 41 | 金 | 21 | 黑 | 27 | 金 | 23 | 玄 | 32 | 寒 | 22 | 清 | 17 | 朱 | 7 | 赤 | 7 |

## 4.2 More Colors in CTP

It is certainly possible to focus on one-character color words as well. We can check the positions of the colors in sentences, and find pairs of colors that appear at the same corresponding positions in a pair of sentences. As the most common color in CTP, “白” corresponds to many other colors: “朱”(zhu1), “丹”(dan1), “紅” (hong2), “緋”(fei1), “彤”(tong2), “青”(qing1), “翠”(cui4), “碧”(bi4), “綠”(lu4), “蒼”(cang1), “清”(qing1), “紫”(zi3), “玄”(xuan2), “皂”(zao4), “黑”(hei1), “淥”(lu4), “明”(ming2), “黃” (huang2), “金”(jin1), and “銀”(yin2).

Table 8 lists some of the frequent color pairs for 10 colors in 10 major columns, each separated by a double bar. In each major column, the **C** sub-column lists the colors that correspond to the color of the major column, and the **F** sub-column shows the frequencies.

That “白” corresponds to “青” (qing1, blue) and “黃” (huang2, yellow) and that “碧” (bi4, green) corresponds to “紅” (hong2, red) and “青” most frequently are s interesting findings.

Given these statistics and other computational supports, we are ready to explore more interesting topics that are related colors in CTP (Cheng et al., 2015).

## 5 Social Network Analysis

Poets mentioned names of their friends or other people in the titles and contents of their poems, so we can use the CTP as a basis for studying social networks of Tang poets. As an extreme example, Li Bai mentioned himself in his own poems: “李白乘舟將欲行，忽聞岸上踏歌聲”[24] (li3 bai2 cheng2 zhou1 jiang1 yu4 xing2, hu1 wen2 an4 shang4 ta4 ge1 sheng1).

In CTP, at least eight poets mentioned Li Bai in 15 works, among which Du Fu contributed seven. We can also see comments on Du Fu by Luo Yin (羅隱), i.e., “杜甫詩中韋曲花，至今無賴尚豪家”[25] (du4 fu3 sh1i zhong1 wei3 qu3 hua1, zhi4 jin1 wu2 lai4 shang4 hao2 jia1).

Of course, mentioning a person’s name may not imply direct friendship. The title “長沙過賈誼宅” (chang1 sha1 guo4 jia3 yi2 zhai2) cannot be used to infer that Liu Zhang-Qing (劉長卿), the author, passed Jia Yi’s (賈誼) home, which is almost impossible as Jia passed away in 168 BC, and Liu was born in 709 AD.

It is easy to build the relationship of “mentioning the name of” in poems, but it takes more discretion to judge direct friendships. We can employ other biographical information such as style names (字, zi4), pen names (號, hao4), birthdays of the poets, from the China Biographical Database[26] to make reliable decisions.

---

[24] source: Li Bai (李白): 贈汪倫 (zeng4 wang1 lun2)

[25] source: 寄南城韋逸人 (ji4 nan2 cheng2 wei3 yi4 ren2)

[26] http://isites.harvard.edu/icb/icb.do?keyword=k35201

Sometimes, a poet's name is not completely listed in poems. In "白也詩無敵，飄然思不群"[27] (bai2 ye3 shi1 wu2 di2, piao1 ran2 si1 bu4 qun2), Du Fu referred to Li Bai only by "白". Hence, cares are needed to handle special cases.

Verbs can offer extra information about the relationships between the poets and the mentioned persons. For instance, Li Shi-Min (李世民) was a Tang emperor, and he "賜" (ci4, give as a present) his poems to officers. Du Fu would "憶" (yi4, recall) Li Bai. Such verbs show us not only the way to find connections between persons but also the types of connections. Poems of emperors, for instance, shed light on their connections with high-ranking officers that are useful for historical studies (cf. Chen, 2010).

We may request a list of such special verbs from domain experts, or we may apply the technique of "word clippers" (Chang, 2006) to find verbs that collocated with names, thus providing opportunities for finding diverse, realistic and virtual connections among poets.

## 6 Concluding Remarks

Finding a needle in a haystack is challenging for human beings, but finding specific words in millions of words is easy for computers. With the aforementioned applications, we demonstrated the potentials of computational tools for studying the Complete Tang Poems, which is a bright spot in a fast-growing research field – Digital Humanities. Computational tools, such as information retrieval, textual analysis, and text mining, cannot accomplish deep research yet, but they can help researchers find and collect much more relevant research material with astonishing efficiency.

Evidence shows that knowing the collection of words that were used by individual poets opens a window for observing the inner worlds of the poets. The concept of distributional semantics is proved to be effective for studying CTP.

We still need to strengthen our ability to check the constraints for pronunciations and rhymes in poems so that we can judge antithesis more precisely with less human participation.

The functions of colors in poems offer a stimulating direction that we intend to dig in further. To do so, we need to employ more technologies for affective computing (Zheng, 2012) so that our software can learn to read between the lines.

We are grateful to the reviewers of this paper for valuable pointers for collocation networks (Williams, 1998) and style analysis (Quiniou et al., 2012). We will have to consider these suggestions in the context of CTP, which contains limited material for individual poets. The actual work about building social networks with CTP is still underway.

### Acknowledgements

This work was supported in part by the Ministry of Science and Technology of Taiwan under grants MOST-102-2420-H-004-054-MY2 and MOST-104-2221-E-004-005-MY3.

## References

Chang, Shan-Pin (張尚斌). 2006. *A Word-Clip algorithm for Named Entity Recognition - by example of historical documents*, Master's thesis, National Taiwan University, Taiwan. (in Chinese)

Chang, Ru-Yng, Chu-Ren Huang, Fengju Lo, and Sueming Chang. 2005. From general ontology to specialized ontology: A study based on a single author historical corpus, *Proc. of the Workshop on Ontologies and Lexical Resources*, 16–21.

Chen, Jack Wei. 2010. *The Poetics of Sovereignty: On Emperor Taizong of the Tang Dynasty*, Harvard University Asia Center, 2010.

Cheng, Wen-Hui, Chao-Lin Liu, Chu-Ting Hsu, and Wei-Yuan Chiu. 2015. Sentiment phenomenology and color politics: Observations of white words in mid-Tang poems, under review.

Chien, Lee-Feng. 1997. PAT-tree-based keyword extraction for Chinese information retrieval, *Proc. of the 20th Annual Int'l ACM SIGIR Conf. on Research and Development in Information Retrieval*, 50–58.

[27] source: Du Fu (杜甫): 春日憶李白 (chun1 ri4 yi4 li3 bai2)

Fang, Alex Chengyu, Fengju Lo, and Cheuk Kit Chinn. 2009. Adapting NLP and corpus analysis techniques to structured imagery analysis in classical Chinese poetry, *Proc. of the Workshop on Adaptation of Language Resource and Technology to New Domains*, 27–34.

Firth, John Rupert. 1957. *A synopsis of linguistic theory 1930–1955*, Studies in Linguistic Analysis, 1–32.

Harris, Zellig. 1954. Distributional structure, *Word*, 10(2-3):1456–1162.

Hu, Junfeng (胡俊峰) and Shiwen Yu (俞士汶). 2001. The computer aided research work of Chinese ancient poems, *ACTA Scientiarum Naturalium Universitatis Pekinensis*, 37(5):725–733. (in Chinese)

Huang, Chu-Ren. 2004. Text-based construction and comparison of domain ontology: A study based on classical poetry, *Proc. of the 18th Pacific Asia Conf. on Language, Information and Computation*, 17–20.

Jiang, Long and Ming Zhou. 2008. Generating Chinese couplets using a statistical MT approach, *Proc. of the 22nd Int'l Conf. on Computational Linguistics*, 377–384.

Jiang, Shao-Yu (蔣紹愚). 2003. "Moon" and "Wind" in Bai Li's and Fu Du's poems – Using computers for studying classical poems, *Proc. of the 1st Int'l Conf. on Literature and Information Technologies*. (in Chinese)

Lee, John. 2012. A classical Chinese corpus with nested part-of-speech tags, *Proc. of the 6th EACL Workshop on Language Technology for Cultural Heritage, Social Sciences, and Humanities*, 75–84.

Lee, John, Ying Cheuk Hui, and Yin Hei Kong. 2013. Treebanking for data-driven research in the classroom, *Proc. of the 4th Workshop on Teaching Natural Language Processing*, 56–60.

Lee, John and Yin Hei Kong. 2012. A dependency treebank of classical Chinese poems, *Proc. of the 2012 Conf. of the North Chapter of the Association for Computational Linguistics: Human Language Technologies*, 191–199.

Lee, John and Tak-sum Wong. 2012. Glimpses of ancient China from classical Chinese poems, *Proc. of the 24th Int'l Conf. on Computational Linguistics*, posters, 621–632.

Lee, Wei-Chih (李瑋質). 2009. *Wen Ting-Yun and Li Shan-Yin's works in the late Tang receive to the Gong-Ti Poetry of the Southern Dynasties*, Master's thesis, National Central University, Taiwan. (in Chinese)

Lin, Dekang. 1998. Automatic retrieval and clustering of similar words. *Proc. of the 36th Annual Meeting of the Association for Computational Linguistics*, 768–774.

Liu, Chao-Lin, Chun-Ning Chang, Chu-Ting Hsu, Wen-Huei Cheng, Hongsu Wang, and Wei-Yuan Chiu. 2015. Textual Analysis of Complete Tang Poems for Discoveries and Applications — Style, Antitheses, Social Networks, and Couplets, *Proc. of the 27th Conf. on Computational Linguistics and Speech Analysis*. (in Chinese)

Lo, Fengju (羅鳳珠). 2005. Design and applications of systems for word segmentation and sense classification for Chinese poems, *Proc. of the 4th Conference on Technologies for Digital Archives*. (in Chinese)

Lo, Fengju. 2008. The research of building a semantic category system based on the language characteristic of Chinese poetry, *Proc. of the 9th Cross-Strait Symposium on Library Information Science*. (in Chinese)

Lo, Fengju, Yuanping Li (李元萍), and Weizheng Cao (曹偉政). 1997. A realization of computer aided support environment for studying classical Chinese poetry, *J. of Chinese Information Processing*, 1: 27–36. (in Chinese)

Miller, George and Walter Charles. 1991. Contextual correlates of semantic similarity, *Language and Cognitive Processes*, 6:1–28.

Quiniou, Solen, Peggy Cellier, Thierry Charnois, Dominique Legallois. 2012. What about sequential data mining techniques to identify linguistic patterns for stylistics? *Proc. of the 13th Int'l Conf. on Intelligent Text Processing and Computational Linguistics*, 166–177.

Voigt, Rob and Dan Jurafsky. 2013. Tradition and modernity in 20th century Chinese poetry, *Proc. of the 2nd Workshop on Computational Linguistics for Literature*, 17–22.

Williams, Geoffrey. 1998. Collocational networks: Interlocking patterns of lexis in a corpus of plant biology research articles

Zheng, Yongxiao (鄭永曉). 2012. Affective computing applied in Chinese classical poetry, *E-Science: Technology & Application*, 3(4):59–66. (in Chinese)

Zhou, Ming, Long Jiang, and Jing He. 2009. Generating Chinese couplets and quatrain using a statistical approach, *Proc. of the 23rd Pacific Asia Conf. on Language, Information and Computation*, 43–52